\documentclass{article}

\usepackage{arxiv}

\usepackage[utf8]{inputenc} 
\usepackage[T1]{fontenc}    
\usepackage{hyperref}       
\usepackage{url}            
\usepackage{booktabs}       
\usepackage{amsfonts}       
\usepackage{nicefrac}       
\usepackage{microtype}      
\usepackage{lipsum}
\usepackage{graphicx}
\graphicspath{ {./images/} }

\usepackage{biblatex}
\bibliography{references.bib}

\title{Location, location, location: Satellite image-based real-estate appraisal \thanks{This abstract was presented at Sixteenth Symposium on Statistical Challenges in Electronic Commerce Research, 2020}} 

\author{
 Jan-Peter Kucklick \\
  Paderborn University\\
  33098, Paderborn \\
  Germany \\
  \texttt{jan.kucklick@uni-paderborn.de} \\
   \And
 Oliver Müller \\
 Paderborn University\\
  33098, Paderborn \\
  Germany \\
 \texttt{oliver.mueller@uni-paderborn.de} \\
}

\begin{document}
\maketitle

Buying a home is one of the most important buying decisions people have to make in their life. Although it is rare that the complete transaction happens online, real estate websites play an increasingly prominent role in the overall decision-making process. For example, the online real estate company Zillow hosts a database of more than 110 million homes in the U.S. and had 196 million unique users in 2019 \cite{zillow}. Besides search and recommendation capabilities, today's real estate websites offer various decision aids to support buyers as well as professional and non-professional sellers in estimating the price of a property. According to Zillow, its home valuation model Zestimate® uses statistical and machine learning models to predict the market value of a property and the prediction "is within 10\% of the final sale price more than 95\% of the time" \cite{zillow_2}.

From a theoretical perspective, real estate evaluation models are based on hedonic pricing models introduced by \citeauthor{lancaster1966new} and \citeauthor{rosen1974hedonic} in the 1970s and 80s \cite{lancaster1966new, rosen1974hedonic}. The main idea of hedonic pricing models is that the overall price of a good can be estimated by decomposing it into its constituting characteristics, determining the value contribution of each individual characteristic, and summing these contributions up to obtain the final price. In the past, most researchers implemented hedonic real estate pricing models through linear regression models \cite{limsombunchai2004house, kok2017big}. While linear models have the advantage of being transparent and interpretable, they typically suffer from high bias, as their assumptions, such as linearity and additivity, often do not align with reality \cite{hastie_09_elements-of.statistical-learning}. 

More recently, driven by technical advances in machine learning algorithms, researchers started to model real estate prices using deep learning models, which are based on deep neural networks that are able to automatically model non-linear relationships and interaction effects \cite{law2019}. The latest research focuses on incorporating new data sources into the modeling process, going beyond structured data. Many attributes concerning a home, such as property type, square footage, or number of bathrooms can easily be measured using numerical and categorical variables (tabular data). However, the decision to buy a home is much more subtle. Real estate buyers are also considering the style and look of the property as well as the appearance and structure of the neighborhood \cite{law2019}. This information is often implicit and, therefore, hard to represent in tabular form. Hence, a small number of researchers has started to create deep learning models, especially convolutional neural networks (CNN), incorporating various types of images of the home and its surrounding into their models in order to capture this information. For example, \citeauthor{poursaeed2018vision} \cite{poursaeed2018vision} evaluated interior images, while \citeauthor{bency2017beyond} \cite{bency2017beyond} and \citeauthor{law2019} \cite{law2019} are using satellite or streetside images in their models. In all studies, a performance increase compared to linear regression baselines without image data was measured.

However, deep learning models, although they are more accurate in prediction \cite{limsombunchai2004house, law2019}, have often been criticized for being intransparent black boxes \cite{lime, montavon2018methods, zeiler2014visualizing}. Interpretability methods aim to demystify the intransparent prediction process of machine learning models by adding human-understandable explanations to model outputs in order to increase the user's trust in the model \cite{lime, du2019techniques}. In most cases when no intrinsic interpretability can be created, a separate model is trained post-hoc to increase transparency \cite{du2019techniques}. Considering the high stakes that are at risk in real estate buying processes, both for buyers and sellers, we argue that interpretability of price estimates is especially important. This is in line with the call for future research on visual explanation of CNNs for real estate valuation by \citeauthor{law2019} \cite{law2019}.

\begin{figure}[ht]
    \begin{center}
        \includegraphics[width=\textwidth]{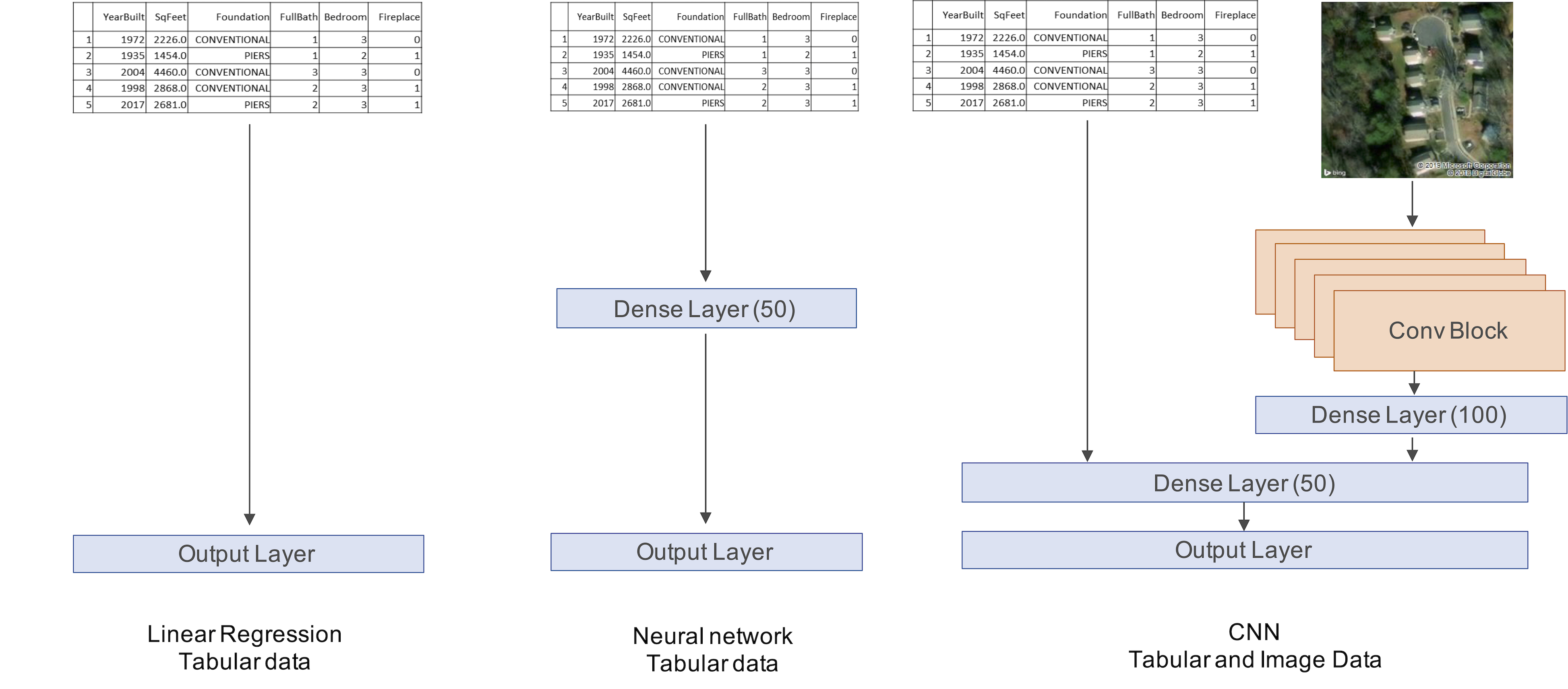}
        \caption{\label{fig:Architecture} Architectures of the different models} 
        The linear regression model represents the standard hedonic model from the literature. The neural network on tabular data is trained as a more sophisticated baseline model for a better comparison and evaluation of the satellite image based model. The CNN trained on tabular and image data is the model we are proposing.
    \end{center}{}
\end{figure}

Against this background, we investigated in how far the inclusion of satellite images improves the predictive accuracy of real estate pricing models and how one can explain the predictions of these models by identifying discriminative visual features between high and low price houses. For our proof-of-concept, we use real estate data from Asheville, North Carolina \cite{asheville}. The Bing Maps API is used to obtain satellite images \cite{Bing} with zoom level 16, depicting 600 by 600 meters around the real estate property. We trained multiple CNNs containing tabular data as well as image data as inputs and the observed house price as the output (see Figure \ref{fig:Architecture}). 

The results show that image data improves the prediction performance of house pricing models (see Table \ref{tab:results}). Compared to a standard hedonic regression model, the use of CNNs with additional image data decreased the mean absolute error (MAE) by 15\%. The more sophisticated baseline is a neural network trained only on tabular data. Comparing the performance of the CNN to this benchmark, we find an 7\% decrease in MAE. Both baseline models are trained on a tabular data set contained numerous information about the property, including, but not limited to, different measures of size, quality, facilities, technical details, and location. 

\begin{figure}[ht]
        \includegraphics[width=\textwidth]{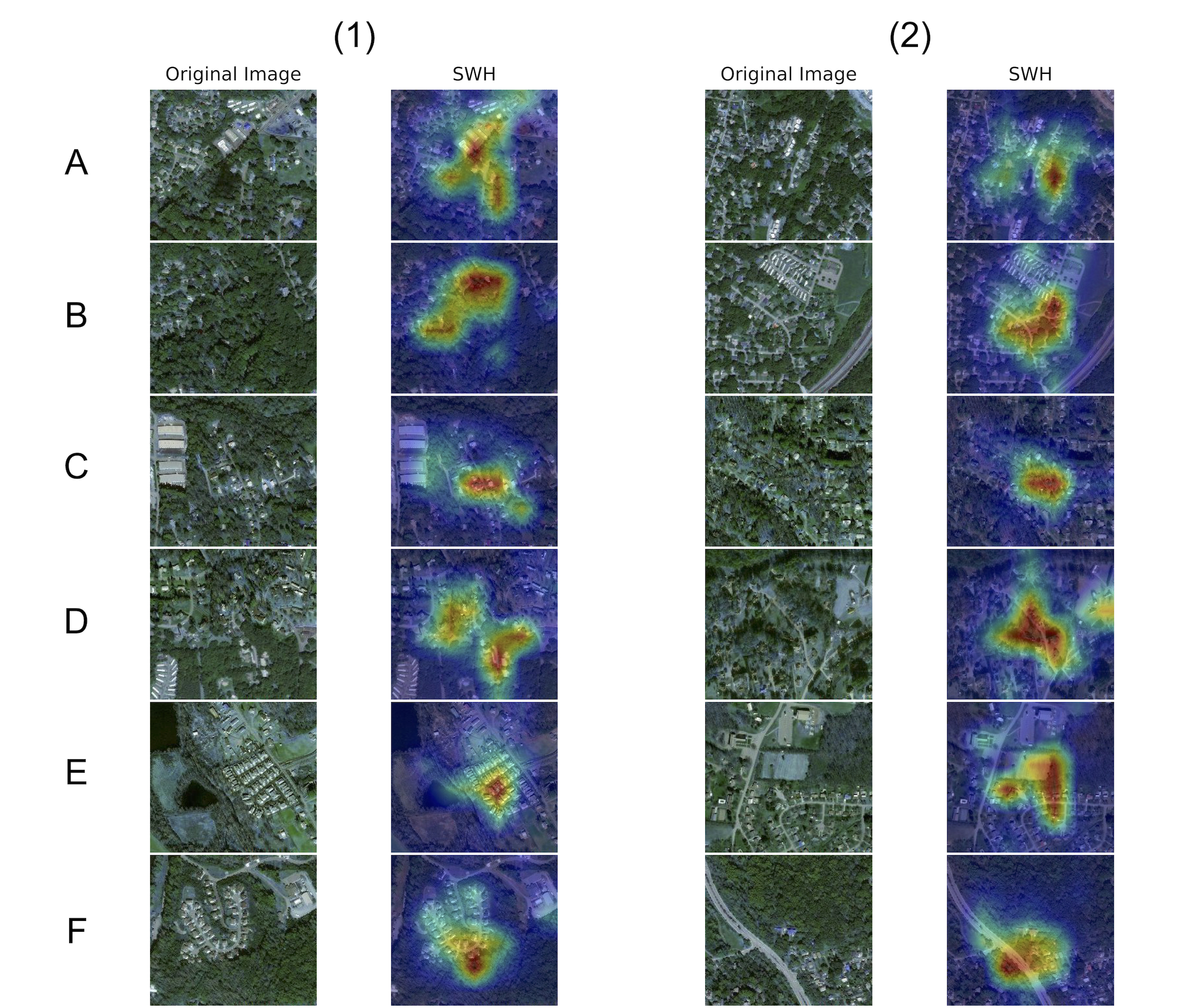}
        \caption{\label{fig:activation_maps} Sliding window heatmap activations on satellite images}
        \begin{center}
            Blue areas highlight regions with no to little influence on the house price, while red regions describe important regions for the price estimation
            
            Images in group 1 depict examples, where SWH works well. Group 2 shows examples, where no clear pattern can be extracted
        
         \textsuperscript{\textcopyright} Microsoft 2019 for the original images
        \end{center}{}
\end{figure}

To make first steps towards explainability of our model, we apply a sliding-window-heatmap (SWH) approach, a visual interpretability method for image-based regression tasks. SWH highlights regions of the image which have a strong influence on the target variable \cite{yang2018ai, zeiler2014visualizing}. Overall, we find that adding SWH on top of the CNN increases the interpretability of the model outputs (Figure \ref{fig:activation_maps}). For example, in Images 1 E and F the residential area is clearly highlighted. In Images 1A and 2A, the building density is partly recognized. In contrast, no interpretable pattern is highlighted in Image 2 F, maybe because the property is located very remotely. When comparing the heatmaps, one has to keep in mind that SWH is a local explainability method, thus the heatmaps of are not comparable across images and patterns might vary strongly between images \cite{ribeiro2016model}. Furthermore, SWH can only detect if the CNN activates on a specific image part, it is not able to determine if the activated region has a positive or negative influence on the response variable (i.e., price). 

\par
During the research process, we encountered the following statistical challenges: In the literature, training multi-input neural networks is described as challenging due to slow convergence and instable learning \cite{bency2017beyond}. Consequently, \citeauthor{bency2017beyond} \cite{bency2017beyond} and \citeauthor{poursaeed2018vision} \cite{poursaeed2018vision} used a two-stage modelling approach separating training of the image regression and tabular regression models. In contrast, \citeauthor{law2019} \cite{law2019} used a one-stage model training the network simultaneously on image and tabular data. We followed the latter approach to allow for interaction effects between the image and tabular features. However, as there is currently no standard architecture for combining image and tabular data in one network, many design decisions regarding how to combine the image and tabular data branches of the network in an effective and efficient way remain to be explored.

In addition, \citeauthor{law2019} \cite{law2019} stated the necessity of visual interpretability methods for evaluating the deep learning model in detail and to overcome its black box characteristics. Most visual model-agnostic interpretability techniques focus on classification instead of regression \cite{lime, montavon2018methods, selvaraju2017grad}. Two methods are currently available for regression interpretability, namely regression activation maps (RAM) \cite{wang2017diabetic} and SWH \cite{yang2018ai}. We use the latter, due to it's architectural independence. Although the visualizations using SWH are not perfect yet and visual model-agnostic explanations are an emerging field of research, we show that the algorithm learns neighborhood structures. Future research should focus on developing better interpretability techniques for image regressions, for example, by coloring image regions according to their positive or negative influence on the response variable.

\par
Concluding, this research aims to improve real estate valuation by using deep learning models that can incorporate image data into the prediction process and increase interpretability of the predictions by applying visual model-agnostic explainability techniques. Online real estate platforms can benefit in two ways from this approach: First, the price estimation is more accurate, leading to a more informed buying decision of the user. Second, as interpretability methods for image regression tasks mature, the user's trust in the estimation might be increased. Both effects should, in the long run, increase the usage of online real estate platforms. However, there are still many open areas for future research on model architecture, variable selection, image zoom level, choice of interpretability technique, and user acceptance of interpretability techniques.

\newpage
\printbibliography

\newpage
\section*{Appendix}

\begin{table}[h]
\begin{center}
\begin{tabular}{|c|c|c|c|}
\hline
\textbf{Model} & \textbf{RMSE} & \textbf{MAE} & \textbf{MAPE} \\ \hline
Linear regression on tabular data & 59,922 & 39,352 & 14.21   \\ \hline
Neural network on tabular data & 55,944 & 35,919 & 12.89 \\ \hline
CNN on tabular \& image data & 51,814 & 33,326 & 12.00  \\ \hline
\end{tabular}
\caption{\label{tab:results} Predictive accuracy of the models}
\end{center}
\end{table}

\end{document}